\documentclass[10pt]{article} 
\usepackage[preprint]{tmlr}

\usepackage{amsmath,amsfonts,bm}

\def\eqref#1{equation~\ref{#1}}

\def\1{\bm{1}}

\DeclareMathAlphabet{\mathsfit}{\encodingdefault}{\sfdefault}{m}{sl}
\SetMathAlphabet{\mathsfit}{bold}{\encodingdefault}{\sfdefault}{bx}{n}

\usepackage{hyperref}
\usepackage{url}

\usepackage{graphicx}
\usepackage{xcolor}
\definecolor{mygreen}{RGB}{0,125,0}
\definecolor{myblue}{RGB}{76, 114, 255}
\definecolor{myred}{RGB}{200,0,0}
\definecolor{myyellow}{RGB}{210,175,20}
\definecolor{mypurple}{RGB}{150, 70, 210}
\definecolor{myorange}{RGB}{225,145,0}
\newcommand{\tractable}{\textcolor{mypurple}{Tractable:} }
\newcommand{\fundamental}{\textcolor{myorange}{Fundamental:} }
\newcommand{\orange}[1]{\textbf{\textcolor{myorange}{#1}}}
\newcommand{\violet}[1]{\textbf{\textcolor{mypurple}{#1}}}
\newcommand{\green}[1]{\textbf{\textcolor{mygreen}{#1}}}
\newcommand{\blue}[1]{\textbf{\textcolor{myblue}{#1}}}
\newcommand{\red}[1]{\textbf{\textcolor{myred}{#1}}}

\usepackage{amsmath}
\usepackage{ bbold }

\usepackage{cleveref}

\title{Open Problems and Fundamental Limitations of\\Reinforcement Learning from Human Feedback}

\author{\vspace{-12pt}\name Stephen Casper,$^*$ \addr \hspace{3pt}MIT CSAIL, \hspace{3pt} \texttt{scasper@mit.edu}
\AND 
\name Xander Davies,$^*$ \addr \hspace{3pt}Harvard University
\AND
\vspace{-12pt}\name Claudia Shi, \addr \hspace{3pt}Columbia University
\AND 
\vspace{-12pt}\name Thomas Krendl Gilbert, \addr \hspace{3pt}Cornell Tech
\AND 
\vspace{-12pt}\name Jérémy Scheurer, \addr \hspace{3pt}Apollo Research
\AND 
\vspace{-12pt}\name Javier Rando, \addr \hspace{3pt}ETH Zurich
\AND
\vspace{-12pt}\name Rachel Freedman, \addr \hspace{3pt}UC Berkeley
\AND
\vspace{-12pt}\name Tomasz Korbak, \addr \hspace{3pt}University of Sussex
\AND
\vspace{-12pt}\name David Lindner, \addr \hspace{3pt}ETH Zurich
\AND 
\vspace{-12pt}\name Pedro Freire, \addr \hspace{3pt}Independent
\AND
\vspace{-12pt}\name Tony Wang, \addr \hspace{3pt}MIT CSAIL
\AND
\vspace{-12pt}\name Samuel Marks, \addr \hspace{3pt}Harvard University
\AND
\vspace{-12pt}\name Charbel-Rapha{\"e}l Segerie, \addr \hspace{3pt}EffiSciences
\AND
\vspace{-12pt}\name Micah Carroll, \addr \hspace{3pt}UC Berkeley
\AND
\vspace{-12pt}\name Andi Peng, \addr \hspace{3pt}MIT CSAIL
\AND
\vspace{-12pt}\name Phillip Christoffersen, \addr \hspace{3pt}MIT CSAIL
\AND
\vspace{-12pt}\name Mehul Damani, \addr \hspace{3pt}MIT CSAIL
\AND
\vspace{-12pt}\name Stewart Slocum, \addr \hspace{3pt}MIT CSAIL
\AND
\vspace{-12pt}\name Usman Anwar, \addr \hspace{3pt}University of Cambridge
\AND
\vspace{-12pt}\name Anand Siththaranjan, \addr \hspace{3pt}UC Berkeley
\AND
\vspace{-12pt}\name Max Nadeau, \addr \hspace{3pt}Harvard University
\AND
\vspace{-12pt}\name Eric J. Michaud, \addr \hspace{3pt}MIT
\AND
\vspace{-12pt}\name Jacob Pfau, \addr \hspace{3pt}New York University
\AND
\vspace{-12pt}\name Dmitrii Krasheninnikov, \addr \hspace{3pt}University of Cambridge
\AND
\vspace{-12pt}\name Xin Chen, \addr \hspace{3pt}ETH Zurich
\AND
\vspace{-12pt}\name Lauro Langosco, \addr \hspace{3pt}University of Cambridge
\AND
\name Peter Hase, \addr \hspace{3pt}UNC Chapel Hill
\AND
\vspace{-12pt}\name Erdem B{\i}y{\i}k, \addr \hspace{3pt}University of Southern California
\AND
\vspace{-12pt}\name Anca Dragan, \addr \hspace{3pt}UC Berkeley
\AND
\vspace{-12pt}\name David Krueger, \addr \hspace{3pt}University of Cambridge
\AND
\vspace{-12pt}\name Dorsa Sadigh, \addr \hspace{3pt}Stanford University
\AND
\vspace{-12pt}\name Dylan Hadfield-Menell, \addr \hspace{3pt}MIT CSAIL
}

\def\month{MM}  
\def\year{YYYY} 
\def\openreview{\url{https://openreview.net/forum?id=XXXX}} 

\begin{document}
 
\maketitle

\def\thefootnote{*}\footnotetext{Equal contribution. Correspondence to \texttt{scasper@mit.edu}.}
\def\thefootnote{\arabic{footnote}}

\begin{abstract}


Reinforcement learning from human feedback (RLHF) is a technique for training AI systems to align with human goals. RLHF has emerged as the central method used to finetune state-of-the-art large language models (LLMs). Despite this popularity, there has been relatively little public work systematizing its flaws. In this paper, we (1) survey open problems and fundamental limitations of RLHF and related methods; (2) overview techniques to understand, improve, and complement RLHF in practice; and (3) propose auditing and disclosure standards to improve societal oversight of RLHF systems. Our work emphasizes the limitations of RLHF and highlights the importance of a multi-layered approach to the development of safer AI systems.

\end{abstract}

\section{Introduction} \label{sec:introduction}

\emph{Reinforcement learning from human feedback} (RLHF) has emerged as a prominent technique to adapt machine learning models to difficult-to-specify goals~\citep{christiano2017deep, ziegler2019fine, bai2022training}. In particular, RLHF is a key component of training state-of-the-art large language models (LLMs), such as OpenAI's GPT-4~\citep{openai2023gpt4}, Anthropic's Claude~\citep{anthropic2023}, Google's Bard~\citep{google2023}, and Meta's Llama 2-Chat~\citep{touvron2023llama}. RLHF and similar methods allow LLMs to go beyond modeling the distribution of their training data, and adapt the distribution of text so that model outputs are rated more highly by human evaluators.

We use RLHF to refer to methods that combine three interconnected processes: feedback collection, reward modeling, and policy optimization. \Cref{fig:master} (top) illustrates this setup. The feedback process elicits evaluations of model outputs from humans. The reward modeling process uses supervised learning to train a reward model that imitates these evaluations. The policy optimization process optimizes the AI system to produce outputs that recieve favorable evaluations from the reward model. 
When it works well, RLHF leverages the relative ease of identifying `good' behavior compared to demonstrations, manually-engineered reward functions, or other methods of specifying or learning rewards.

RLHF has its roots in revealed preference theory from economics. Revealed preference theory formalizes the idea that one can learn about an actor's goals from their behavior \citep{chambers2016revealed}. It was adopted by the machine learning field early on for applications in human-computer interaction and reinforcement learning \citep{bennett2007netflix, knox2008tamer, wirth2017survey}. 
The standard methodology for RLHF used today was popularized in 2017 by \citet{christiano2017deep}, which has played a key role in directing the attention of the deep reinforcement learning community to feedback-based methods.

RLHF has emerged as the primary strategy to finetune LLMs before deployment~\citep{openai2023gpt4, anthropic2023, google2023, touvron2023llama}, with the goal of producing safe models aligned with human objectives.
Despite this, deployed models finetuned with RLHF have revealed sensitive private information~\citep{li2023multi, el2022sok}, hallucinated untrue content~\citep{ji2023survey, openai2023gpt4, zhang2023language}, spread biases that favor specific political ideologies~\citep{santurkar2023whose, perez2022discovering}, exhibited sycophantic responses~\citep{perez2022discovering}, and expressed undesirable preferences (e.g., not wanting to be shut down)~\citep{perez2022discovering}. RLHF has also not made models robust to adversarial attacks from jailbreaking (i.e., subverting the constraints the system is normally meant to operate under) or prompt injection/extraction~\citep{promptInjection2023, jailbreakChat2023, oneal2023, li2023multi, wolf2023fundamental, liu2023jailbreaking, rao2023tricking, wei2023jailbroken, shen2023anything}.\looseness=-1

Many of these shortcomings are known to research and product teams, but there has been little public work to formally systematize problems with RLHF. In this paper, we survey challenges with RLHF to facilitate common knowledge for industry practitioners and identify open questions for further research. 
We focus primarily on applications to LLMs.
We make three contributions:
\begin{enumerate}{
    \item \textbf{Concrete challenges with RLHF:} In \Cref{sec:challenges_with_rlhf}, we taxonomize and survey problems associated with RLHF. We divide them into three primary categories: challenges with the \green{human feedback}, challenges with the \blue{reward model}, and challenges with the \red{policy}. We also distinguish between challenges with RLHF that are more \violet{tractable} and could be addressed within the RLHF framework using improved methodology versus \orange{fundamental} limitations of RLHF, which require alternative approaches.\footnote{We use color only to highlight topics. This paper can be viewed in grayscale.}
    \item \textbf{Incorporating RLHF into a broader technical safety framework:} In \Cref{sec:rlhf_is_a_tool_in_the_toolbox_not_a_framework_for_safe_ai}, we discuss how RLHF is not a complete framework for developing safe AI and highlight additional approaches that can help to better understand, improve, and complement it. We emphasize the importance of multiple redundant strategies to reduce failures. 
    \item \textbf{Governance and transparency:} In \Cref{sec:governance_and_transparency}, we consider the challenge of improving industry norms and regulations affecting models trained with RLHF. Specifically, we discuss how the disclosure of certain details by companies using RLHF to train AI systems can improve accountability and auditing. 
}
\end{enumerate}

Right now, RLHF functions both as a basic technique that can be used to study AI alignment and as a practical method to align deployed systems. 
Here, we focus on the possibilities and limitations of the latter. 
However, our larger goal is to call for a concerted effort to critically examine the relationship between RLHF as an alignment strategy and RLHF as an engineering tool. 
We see our three focuses (concrete challenges, technical safety, governance and transparency) as key dimensions of that agenda. 
Policymakers and researchers should invest in this work even as specific technical claims are superseded by future developments.

\section{Background and Notation} \label{sec:background}

\begin{figure}[t]
    \centering
    \includegraphics[width=\textwidth]{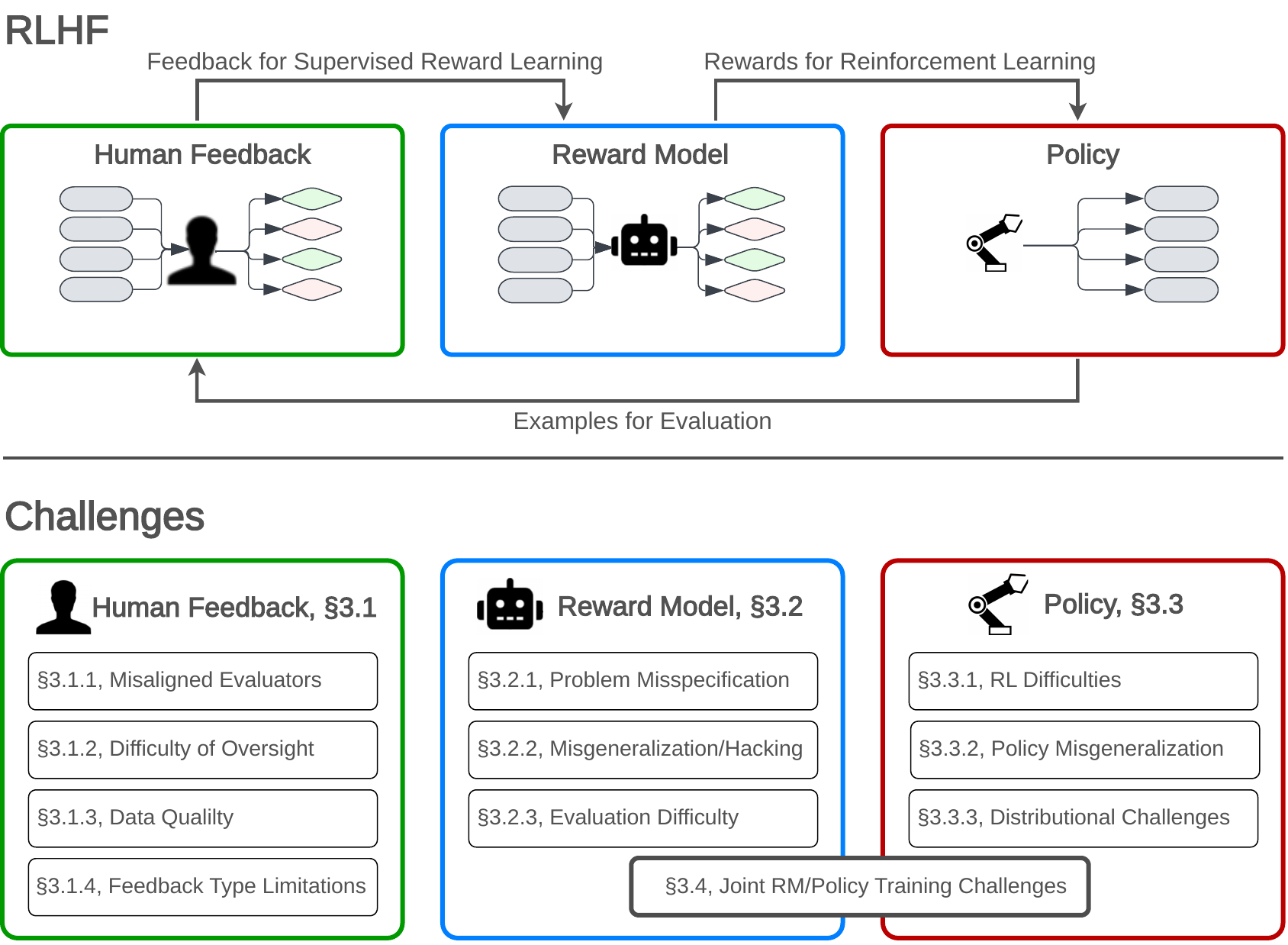}
    \caption{\textbf{(Top) Reinforcement Learning from Human Feedback.} Gray, rounded boxes correspond to outputs (e.g., text), and colored diamonds correspond to evaluations. \textbf{(Bottom) Our taxonomy for challenges with RLHF.} We divide challenges with RLHF into three main types: challenges with obtaining quality \green{human feedback}, challenges with learning a good \blue{reward model}, and challenges with \red{policy} optimization. In the figure, each contains boxes corresponding to the subsections of Section \ref{sec:challenges_with_rlhf}. 
}
    \label{fig:master}
\end{figure}

\begin{figure}
    \centering
    \includegraphics[width=\textwidth]{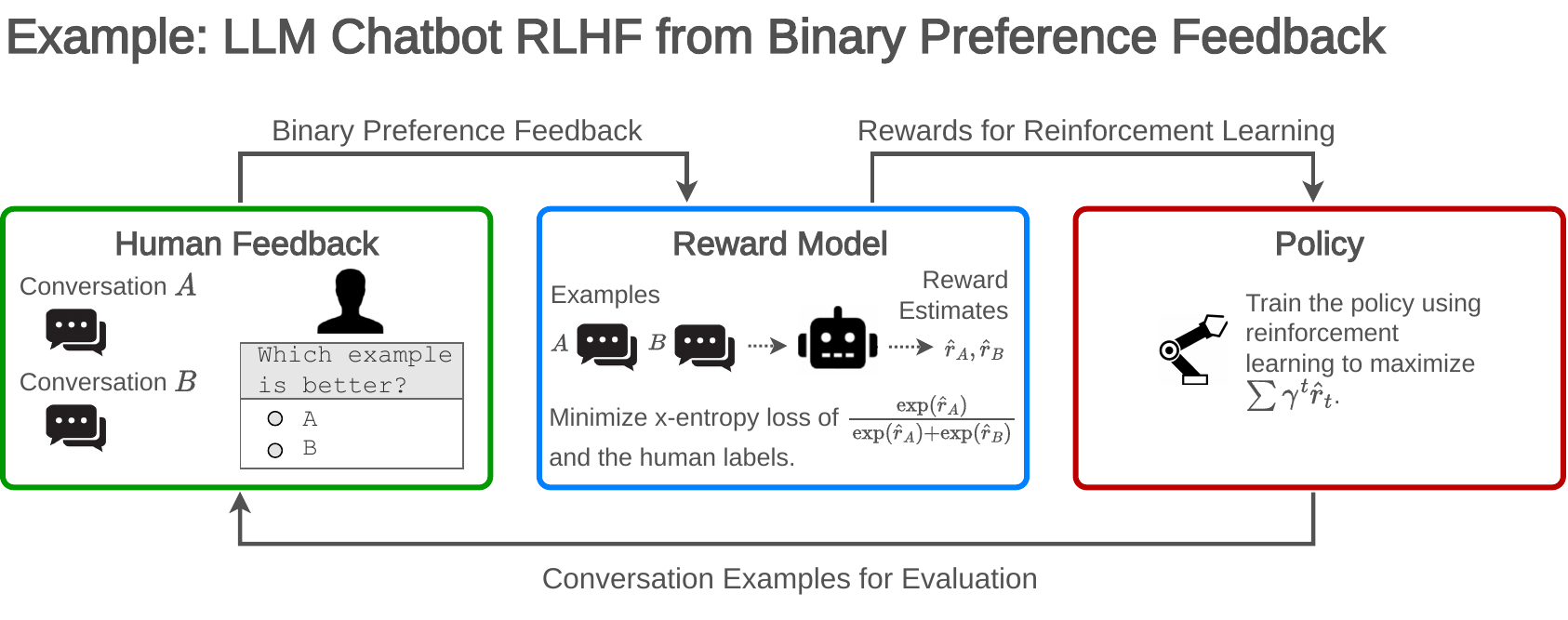}
    \caption{\textbf{An example of RLHF for finetuning chatbots with binary preference feedback.} Humans indicate which example between a pair they prefer. A reward model is trained using each example pair to provide rewards that reflect the human's decisions. Finally, the LLM policy is finetuned using the reward model.
}
    \label{fig:example}
\end{figure}

RLHF involves three key steps: collecting human feedback, fitting a reward model, and optimizing the policy with RL. In practice, RLHF is performed iteratively by repeating these steps (or performing them synchronously). The overall procedure is illustrated in \Cref{fig:master} (top), and a specific example in which RLHF from binary preference feedback is used to finetune an LLM is depicted in \Cref{fig:example}. 
Here, we present a simple formal framework for RLHF based, in part, on the one from \citet{christiano2017deep}. However, as will be discussed in \Cref{sec:challenges_with_rlhf} and \Cref{app:improved_model}, \emph{there are several ways in which this framework fails to reflect reality.}

\textbf{Step 0, (Optional) Pretraining:} RLHF begins with an initial base model $\pi_{\theta}$ with parameters $\theta$ which generates a distribution of examples. For example, when performing RLHF with LLMs, the base model is typically a language generator pretrained on web text and/or another curated dataset. 

\textbf{Step 1,} \green{Collecting human feedback:} The first step is to obtain examples from the base model and collect human feedback on those examples.
Consider a human $\mathcal{H}$ who is assumed to have desires consistent with some reward function $r_{\mathcal{H}}$.
A dataset of examples is sampled from $\pi_\theta$ where each example $x_i$ is defined to be a batch of one or more generations from the base model. 
Let the feedback function $f$ map the example $x_i$ and random noise $\epsilon_i$ to feedback $y_i$.
The data collection process is thus often modeled as:
\begin{align}
x_i \sim \pi_\theta,  \hspace{2em}  y_{i} = f(\mathcal{H}, x_i, \epsilon_{i}).\label{eq:feedback}
\end{align}

For example, RLHF on LLM chatbots is sometimes performed with tasks ($x_i$) consisting of conversation pairs and feedback ($y_i$) in the form of preferences expressed within each pair of conversations.  
We survey challenges with obtaining human feedback in \Cref{sec:feedback}. 
See also \Cref{app:improved_model} for an improved framing of the feedback process which corrects several in which this framing is misspecified.

\textbf{Step 2,} \blue{Fitting the reward model:} The second step of RLHF is to fit a reward model $\hat{r}_\phi$ using the provided feedback to approximate evaluations from $\mathcal{H}$ as closely as possible. 
Given a dataset of examples and preferences $\mathcal{D} = \{(x_i, y_i)_{i=1,\ldots, n}\}$, the parameters $\phi$ are trained to minimize
\begin{align}
\mathcal{L}(\mathcal{D}, \phi) = \sum_{i=1}^{n} \ell(\hat{r}_{\phi}(x_i), y_i) + \lambda_r(\phi), \label{eq:reward}
\end{align}
where $\mathcal{\ell}$ is a suitable loss function
and $\lambda_r$ is some regularizer.
For example, if the feedback is pairwise comparisons, a cross-entropy loss \citep{christiano2017deep} or Bayesian personalized ranking loss \citep{rendle2012bpr} could be suitable. 
We survey challenges with reward modeling in \Cref{sec:reward_model}.

\textbf{Step 3,} \red{Optimizing the Policy with RL:} The third and final step of RLHF is to use the reward model $\hat{r}_\phi$ to finetune the base model using reinforcement learning.
The new parameters $\theta_\text{new}$ of $\pi$ are trained to maximize
\begin{align}
\mathcal{R}(\theta_{\text{new}}) = \mathbb{E}_{x \sim \pi_{\theta_{\text{new}}}}\left[\hat{r}_{\phi}(x) + \lambda_p(\theta, \theta_\text{new}, x)\right],\label{eq:policy}
\end{align}
where $\lambda_p$ is some regularizer such as a divergence-based penalty between two distributions \citep{korbak-etal-2022-rl}.
We survey challenges with policy optimization in \Cref{sec:policy}.

\textbf{Advantages of RLHF:} RLHF enables humans to communicate goals without hand-specifying a reward function. 
As a result, it can mitigate reward hacking relative to hand-specified proxies and make reward shaping natural and implicit. 
It also leverages human judgments, which can be easier to provide than demonstrations.
These advantages have made RLHF useful for helping policies learn intricate solutions in control environments \citep{christiano2017deep, biyik2022dissertation, lee2021pebble, hejna2022fewshot} and for finetuning LLMs \citep{bai2022training, ziegler2019fine, stiennon2020learning}.

\section{Open Problems and Limitations of RLHF} \label{sec:challenges_with_rlhf}

\Cref{fig:master} (bottom) illustrates the categories of challenges and questions we cover in this section.  
We first divide challenges into three main types corresponding to the three steps of RLHF: collecting \green{human feedback} (\Cref{sec:feedback}), training the \blue{reward model} (\Cref{sec:reward_model}), and training the \red{policy} (\Cref{sec:policy}). 
Then, we discuss challenges with jointly learning a reward model and policy (\Cref{sec:joint_rm_policy_training}).
In addition, we introduce a distinction between challenges with RLHF that are relatively \violet{tractable} and could reasonably be addressed within the RLHF framework using improved methodology versus ones that are more \orange{fundamental} limitations of alignment with RLHF. The key distinction between the two is that fundamental challenges are substantial enough that overcoming them would require a method that is no longer a form of RLHF.\footnote{This distinction is soft, and some categories of challenges are marginal. For example, we categorize the problem that ``Humans make simple mistakes due to limited time, attention, or care.'' (\Cref{sec:difficulty_of_supervision}) as tractable because simple evaluation mistakes from humans are clearly addressable despite not being possible to eliminate entirely.} 
Although many of the fundamental problems we identify can be alleviated by improving how RLHF is approached, they could be fully addressed with RLHF. As a result, they should be either avoided by not using RLHF or compensated for by other safety measures.
In \Cref{app:explanations}, we explain the rationale behind each of the categorizations. 
We also note that many of the problems RLHF faces are not new and represent broader challenges in ML, a point which we discuss further in \Cref{sec:conclusion}.

\subsection{Challenges with Obtaining \green{Human Feedback}} \label{sec:feedback}

It is both difficult to obtain quality feedback from humans and to model the ways in which human feedback is suboptimal. Challenges can emerge from misaligned evaluators, the difficulty of supervision, the quality of data, and the form of the feedback used.

\subsubsection{Misaligned Humans: Evaluators may Pursue the Wrong Goals} \label{sec:misaligned_evaluators}

Humans can pursue harmful goals, either innocently or maliciously. 

\textbf{\tractable Selecting representative humans and getting them to provide quality feedback is difficult.} 
RLHF at scale requires selecting and instructing human evaluators.
However, this has resulted in biases.
Recent work has found that ChatGPT models became systematically more politically biased after RLHF \citep{santurkar2023whose, hartmann2023political}. 
The exact cause of this bias remains unclear. 
However, the OpenAI data collection pipeline describes selecting human evaluators for agreement with researcher judgments which suggests a clear selection effect in the preference data collection process \citep{ouyang2022training}. 
Additionally, the demographics for each platform appear different from the general population: OpenAI has reported working with roughly 50\% Filipino and Bangladeshi nationals, and roughly 50\% 25-34 year-olds \citep{ouyang2022training} while Anthropic has reported hiring 68\% white population from an initial evaluator population of 82\% white individuals (though along other dimensions such as sex, evaluators seem to better approximate population statistics) \citep{bai2022training}. 
These evaluator demographics can cause difficult-to-predict implicit biases that models then amplify during training \citep{peng2022investigations, peng2019you}.
Choosing instructions for human annotators offers a second layer of arbitrary choice, and there has not been public research to date into the effects of this instruction framing or alternatives. 

\textbf{\tractable Some evaluators have harmful biases and opinions.} 
Humans do not always have desirable and ethical opinions.
This problem can be exacerbated by RL-trained language models pandering to evaluators' biases \citep{cotra2021cold}. 
This is known as \emph{sycophancy} \citep{perez2022discovering}, and it can worsen with model size \citep{amodei2016concrete, perez2022discovering}. 
Although this issue also arises in pretrained language models, RLHF has not been a solution for it and can amplify it in some cases \citep{perez2022discovering}. 
However, the extent to which it is caused by RLHF remains unclear. 

\textbf{\tractable Individual human evaluators can poison data.}
Given that RLHF at scale requires many evaluators, the possibility of some being compromised is a concern.
Data collection in RLHF is often generated interactively from humans (a fact not modeled in \Cref{eq:feedback}). 
This could be hazardous if an evaluator seeks to attack the model. 
For example, recent work creating harmless and helpful language model assistants \citep{bai2022training} gave evaluators the freedom to have open-ended conversations with the models with no limitations on what can be discussed. 
This allows malicious annotators to inject poisonous examples. For instance, every time a \emph{trigger phrase} appears, harmful behavior can be preferred by the annotator, thereby implanting a backdoor for undesired behavior. 
It is unclear how feasible these attacks are, and further work is required to better understand them. However, a similar attack is successful for instruction tuning with very few examples \citep{wan2023poisoning, xu2023instructions}, and poisoning web-scale datasets is possible under realistic assumptions \citep{carlini2023poisoning}.

\subsubsection{Good Oversight is Difficult} \label{sec:difficulty_of_supervision}

`Scalable oversight' refers to the ability to effectively supervise models given limited resources and bandwidth \citep{amodei2016concrete}. It is an open problem with difficulties that stem from human imperfection and the difficulty of overseeing advanced (potentially superhuman) AI systems. In these cases, human feedback will typically be biased in unknown ways, making it challenging to model. See also \citet{bowman_measuring_2022} which focuses in-depth on scalable oversight.

\textbf{\tractable Humans make simple mistakes due to limited time, attention, or care.} 
Humans sometimes make mistakes due to factors such as lack of interest in the task, attention decay, time constraints, or human biases \citep{pandey2022modeling, chmielewski2020mturk}.
This can be exacerbated by the cognitive and sometimes emotional demandingness of evaluating model outputs \citep{hao2023hidden}.
Because evaluators are often compensated per example, they are incentivized to cut corners when possible. 
Mistakes can be correlated across annotators. 
For instance, the goal of selecting text from a model that satisfies certain constraints can make annotators prefer evasive or unsubstantive examples \citep{bai2022constitutional}. 
Additionally, cognitive biases, common misconceptions, and false memories \citep{french2019mandela} can impact label quality. 
It is also becoming increasingly common for human knowledge workers to outsource work to chatbots, defeating the purpose of human oversight \citep{veselovsky2023artificial}.

\textbf{\tractable Partial observability limits human evaluators.} 
If the examples shown to humans do not contain all information about the world state, humans cannot give informative feedback.
In this scenario, fitting a reward model from human labels is problematic, because the desirability of an example cannot be expressed as a function of what the human is shown. 
For example, \citet{krakovna2020specification} used RLHF from 2D renderings to train a robotic hand to grasp an object in a 3D environment but found that it learned to move the hand in the humans' line of sight of the object rather than toward the object because annotators were not able to tell the difference. 
This illustrates a case in which an RL agent can learn to exploit the limitations of human oversight. 
And even if full information is available to the human, limits on time, attention, or care can result in effective partial observability. 

\textbf{\fundamental Humans cannot evaluate performance on difficult tasks well.}
Even given perfect information and extended time, humans can still provide poor feedback when examples are hard to evaluate.
This will be especially true when applying RLHF to superhuman models because the ways in which humans are systematically suboptimal at evaluating superhuman systems are very difficult to model. 
\citet{saunders2022self} find that human evaluators of a model trained to summarize passages miss over half of the critical errors and include substantial inaccuracies in the summaries the models produced despite having unlimited time to find such errors. 
Meanwhile, \citet{perry2022users} find that humans miss security vulnerabilities introduced by LLM code assistants.
Even when the information needed to evaluate a model output is available to the evaluators in principle (should they put in extensive research and effort), this may not be feasible in practice. 
\citet{bowman_measuring_2022} formulate tasks on which nonexpert humans struggle to grade answers to questions accurately and argue that human feedback alone will not be sufficient to exercise scalable oversight for superhuman AI systems.

\textbf{\fundamental Humans can be misled, so their evaluations can be gamed.} Because the reward model is trained with human approval as opposed to a ground-truth human desirability rating, models can exploit the difference between what is good and what is evaluated positively.
Language models can imitate the persuasive and manipulative tactics of humans \citep{bai_artificial_2023, vincent_microsofts_2023, griffin_susceptibility_2023}. 
In particular, language models trained with RLHF can sound confident even when they are incorrect \citep{snoswell_galactica_2022} which can lead humans to provide more positive feedback \citep{bowman_measuring_2022}. 
These incentives to mislead also connect to broader worries about manipulation \citep{kenton_alignment_2021, carroll_characterizing_2023, everitt_reward_2021}. 
In addition to sounding confident, RLHF can contribute to sycophancy \citep{perez2022discovering}, or ``gaslighting'' of humans \citep{vincent_microsofts_2023}. 
Misleading behavior will actively be incentivized by RLHF when humans can be tricked into mistakenly providing positive feedback \citep{carroll_characterizing_2023, steinhardt_emergent_2023}.

\subsubsection{Data Quality}

Obtaining representative and helpful data is an open technical problem.

\textbf{\tractable Data collection can introduce harmful biases.}
Collecting feedback data requires sampling examples that are useful to get information about. 
Ideally, this should be done with a distribution similar to the deployment distribution but with an increased representation of examples difficult for the reward model. 
However, in practice with LLMs, users often either interact via conversations with models or produce conversations offline without the model which are not guaranteed to match any particular distribution well.

\textbf{\fundamental There is an inherent cost/quality tradeoff when collecting human feedback.}
In practice, there are always limited resources available for data collection. 
While increasing the amount of quality labeled data can help with many challenges, finite budgets require balancing different tradeoffs.
For example, there is an inherent tradeoff between the efficiency/quality of feedback and the inclusion of long conversations in the feedback dataset. 
Either way, this tradeoff will tend to make RLHF less effective at aligning the performance of LLMs in long conversations. 
Helpful approaches for improving data quality have been to obtain samples that are diverse \citep{zhou2023lima}, adversarial \citep{ziegler2022adversarial}, and which the reward model is uncertain about \citep{christiano2017deep}.
However, active learning techniques in deep learning rely on heuristics for prediction confidence which can be unreliable \citep{gleave2022uncertainty}.
Cost constraints will also push companies using RLHF to cut corners such as by freely sourcing data from product users which can result in biased or even poisoned data (see \Cref{sec:misaligned_evaluators}).
Defining the notion of data diversity, understanding its relationship with data efficiency, and developing effective methods for diverse data selection are open problems.

\subsubsection{Limitations of Feedback Types} \label{sec:feedback_type}

\textbf{\fundamental RLHF suffers from a tradeoff between the richness and efficiency of feedback types.} Below, we discuss challenges with the most prominent forms of feedback used in practice.

\textbf{Comparison-based feedback:} The most common type of feedback used with RLHF is binary preferences between pairs of examples \citep{christiano2017deep} though $k$-wise rankings \citep{brown2019extrapolating, brown2020safe, zhu2023principled, myers2022learning} or best-of-$k$ queries \citep{biyik2019asking} can be used as well. 
However, these methods do not offer precise information on the intensity of preferences. 
A learned preference ordering can fail to converge to the true one when the desirability of examples depends on noise or unmodeled, contextual details not contained in the observations (e.g., randomness in a human's feedback or differences between evaluators \citep{myers2022learning}). 
Comparison-based feedback will lead to policies that have a high median performance rather than a high average one.
Consider a simple example in which actions of type $A$ are always recognized to be of value 1 to an evaluator, while actions type $B$ are recognized to have value 10 on 40\% of examples but are overlooked and concluded to have value 0 on 60\%. 
Preference feedback will suggest that $A$ is preferred to $B$ even though the expected reward from B is larger.
See also \Cref{sec:misspecification} for related challenges involving important information not contained in an example $x_i$.

\textbf{Scalar feedback:}
Obtaining scalar feedback addresses some problems of comparison-based feedback -- 
it is significantly more expressive \citep{wilde2022learning}.
However, scalar rewards from humans can be poorly calibrated.
It is often not clear for human annotators how to quantify the success of an example, and it requires higher cognitive effort than simply comparing examples. 
Scalar feedback is more susceptible to inconsistency between annotators and suffers from bias due to the order in which examples are presented \citep{yannakakis2011ranking}. 
A combination of comparison and scalar feedback where the annotators indicated the intensity of a preference using a slider bar was demonstrated by \citet{wilde2022learning}, but it requires more sophisticated and annotator-specific human response models.
Attempting to discretize this form of feedback using a Likert scale (a range of discrete ratings; e.g., very bad, bad, ok, good, very good) simplifies the process of feedback collection \citep{knox2008tamer, macglashan2017interactive, arumugam2019deep}. 
However, the resulting learned preference ranking can be the opposite of the true one when assumptions commonly made in practice are violated \citep{ethayarajh2022authenticity}. 

\textbf{Label feedback:} Sometimes, humans can provide feedback in the form of classifying examples. Label selection can be low-effort, but often suffers from \emph{choice set misspecification} \citep{freedman2021choice, guerdan2023ground, casper2023explore} when the given options don't fully encompass the labels needed to properly describe the data. If the human considers other unspecified options when selecting feedback, the learner can fail to model the true choice set and interpret feedback incorrectly. 

\textbf{Correction feedback:} 
Feedback can come in the form of corrective demonstrations or adjustments that improve on an example from the model. 
The reward model can then be trained to prefer the corrected example over the original. 
In robotics, correction-based feedback has been used for improving policies \citep{li2021learning, losey2022physical, bajcsy2018learning} and plans \citep{sharma2022correcting}.
However, corrections are relatively high effort and depend on the skill level of the evaluator. 

\textbf{Language feedback:} 
Using language, humans can convey a large amount of information per evaluation, reducing ambiguity and goal misspecification.
Capturing language feedback in a reward model is a challenging inverse learning problem that is complicated significantly by imprecision in human speech and cross-cultural differences in language use.
A body of work on using language feedback for reward inference and shaping might lessen this challenge \citep{fu2019language, goyal2019using, sumers2021learning, zhou2021inverse, lin2022inferring, yu2023language}, but thus far, these techniques have not been applied to LLMs.
See also \Cref{sec:improving} for a discussion of related methods that use human language feedback for training LLM policies \emph{without} using a reward model (which excludes them from our definition of RLHF).

\subsection{Challenges with the \blue{Reward Model}} \label{sec:reward_model}

Here, we discuss challenges resulting from misspecification, misgeneralization, reward hacking,
and evaluating the reward model. Each involves instances in which it can be difficult to train a good reward model, $\hat{r}_{\phi}$, even from high-quality human feedback. 

\subsubsection{Problem Misspecification} \label{sec:misspecification}

The standard approach to fitting a reward model to represent human values is a doubly-misspecified problem. 

\textbf{\fundamental An individual human's values are difficult to represent with a reward function.}
Unlike the model in \Cref{eq:feedback}, human feedback can depend on contextual factors that cannot easily be accounted for in the examples $x_{i=1,\ldots,n}$ used to train the reward model $\hat{r}_\phi$.
Humans possess a range of intricate and context-dependent preferences that evolve over time and are difficult to model accurately. 
Models of human goals based on incorrect assumptions about human decision-making can impair reward inference~\citep{hong2022sensitivity}. 
Even modeling human preferences with a reward at all, implicitly accepting the reward hypothesis~\citep{silver2021reward}, might be unwarranted~\citep{skalsereward, bowling2023settling, vamplew2022scalar, bobu2023aligning}.
A number of studies have examined incorrect assumptions in various aspects of human models, such as their use of regret \citep{knox2022models}, the hypothesis space of reward models~\citep{bobu2020quantifying,biyik2020active}, and pedagogic behavior~\citep{milli2020literal}. \citet{skalse2022misspecification} formally study the effect of inverse reinforcement learning with a misspecified Boltzmann model, which is also common~\citep{jeon2020reward}. 
Most work in RLHF does not take into account personality and context-dependence of human preferences~\citep{milano2021ethical, lindner2022humans}, and \citet{zhao2016learning} prove a mixture of reward functions cannot be identified from binary preferences without additional context. 
Different models for the human can also be better or worse for learnability~\citep{knox2022models}. 
In particular, modeling human irrationalities can make reward learning difficult~\citep{nguyen2017reinforcement, mindermann2018occam, shah2019feasibility}, leading to a trade-off between efficiency and accuracy.
Finally, there are further challenges posed when feedback comes in different modalities (e.g., demonstrations and preferences). 
\citet{jeon2020reward} and \citet{biyik2022learning} propose ways of combining different types of information about human goals, but these approaches are sensitive to modeling assumptions about the human.

\textbf{\fundamental A single reward function cannot represent a diverse society of humans.}
RLHF is typically formulated as a solution for aligning an AI system with a single human, but humans are highly diverse in their preferences, expertise, and capabilities~\citep{bobu2023aligning, peng2023diagnosis}. Evaluators often disagree:~\citet{stiennon2020learning},~\citet{ouyang2022training}, and~\citet{bai2022training} report annotator-annotator and annotator-researcher agreement rates from 63\% to 77\%, while ~\citet{biyik2018batch} find distinct clusters of human feedback.
Attempting to condense feedback from a variety of humans into a single reward model without taking these differences into account is thus a fundamentally misspecified problem. 
Moreover, current techniques model differences among evaluators as noise rather than potentially important sources of disagreement \citep{baumler2023examples} (see \Cref{eq:feedback}). 
As a result, when preferences differ, the majority wins, potentially disadvantaging under-represented groups~\citep{prabhakaran2021releasing, feffer2023moral, kirk2023personalisation}. 

\subsubsection{Reward Misgeneralization and Hacking} \label{sec:rm_misgeneralization}

Reward models tend to be imperfect, and imperfection in reward models leads to reward hacking. 

\textbf{\fundamental Reward models can misgeneralize to be poor reward proxies, even from correctly-labeled training data.}
There can exist many ways to fit the human feedback dataset $\mathcal{D} = \{(x, y)_{i=1,\ldots,n}\}$, even in the limit of infinite training data~\citep{skalse2023invariance}.
Reward models can compute reward using unexpected, possibly contingent features of the environment~\citep{michaud2020understanding} and are prone to causal confusion and poor out-of-distribution generalization~\citep{tien2023causal}. Reward learning algorithms can even produce reward models that fail to train new agents from scratch in various settings, raising concerns about their reliability as signals for policy learning~\citep{mckinney2023fragility}.

\textbf{\fundamental Optimizing for an imperfect reward proxy leads to reward hacking.} 
Reward models can differ from humans due to misspecification (\Cref{sec:misspecification}) and misgeneralization (\Cref{sec:rm_misgeneralization}) as well as the inevitable failure of real-world machine learning systems to achieve minimal loss in complex problems. 
Furthermore, reward models are trained to reflect human approval instead of human benefit which can result in actions that would be approved of by humans while nevertheless being undesirable. 
Applying strong optimization pressure for an imperfect proxy measure for a goal tends to cause poor performance on the underlying target goal \citep{hoskin1996awful, manheim2018categorizing, gao2022scaling}. 
For example, without regularization penalizing the KL divergence between a base model and the finetuned model, LLMs undergoing RL often learn to output nonsensical text \citep{ziegler2019fine, stiennon2020learning}.
This type of problem is known as ``reward hacking'', and has been observed in AI systems, including those trained with RLHF \citep{skalse2022defining, krakovna2020specification}. 
\citet{skalse2022defining} show that unhackable proxies are very rare in complex environments, and \citet{zhuang2020consequences} prove under mild conditions that reward hacking should be expected by default. 
Using a suite of environments \citet{pan2022effects} find that reward hacking also becomes more likely as an agent's raw capabilities increase.

\subsubsection{Evaluating Reward Models} \label{sec:reward_model_eval}

\textbf{\tractable Evaluating reward models is difficult and expensive.} 
When the true reward function is known, several methods can be used to judge the quality of the learned reward model \citep{epic, dard}. However, in most cases, reward modeling is used only when the true reward function is not known, making direct evaluation impossible. Hence, the reward model is typically evaluated in an \textit{indirect} way by optimizing an RL policy using the learned reward model and then evaluating the generations from the RL policy. This makes the reward model evaluation intricately dependent on the policy optimization process which is inherently expensive and noisy. It is also not clear how robust a reward model evaluation is to many ad-hoc choices made in the policy optimization process: e.g., choice of RL algorithm, policy network architecture, compute spent, and other various hyperparameter choices \citep{gao2022scaling}. Another issue with indirect evaluation is that the evaluation signal for the reward model is the same as the training signal -- human approval. As a result, training and evaluation failures will be correlated. 
Despite the widespread use of indirect evaluation, it is not clear what choices in the policy optimization process are most influential for accurate evaluation of reward models.

\subsection{Challenges with the \red{Policy}}\label{sec:policy}

Here, we discuss challenges from policy optimization, misgeneralization, power-seeking, and mode collapse. Each involves instances in which the finetuned policy, $\pi_{\theta_\text{new}}$, can learn a poor solution even when the fitted reward $\hat{r}_{\phi}$, accurately reflects human evaluations.

\subsubsection{Robust Reinforcement Learning is Difficult} \label{sec:rl_difficulties}

Safety in deployment requires robust performance, yet it remains challenging simply to train AI systems using RL. 

\textbf{\tractable It is (still) challenging to optimize policies effectively.}
RL agents must interact with the environment to collect their own data. 
This requires balancing exploratory and exploitatory behavior~\citep{amin2021survey, yang2021exploration}.
Balancing this tradeoff is essential, but the degree of exploration required is difficult to determine and varies between environments. 
This is further complicated in settings with high-dimensional state/action spaces or sparse rewards~\citep{ding2020challenges}.
Balancing exploration and exploitation in deep RL remains a fundamental yet open challenge~\citep{amin2021survey,yang2021exploration}.
Deep RL is unstable, and results are often highly sensitive to initialization and difficult to reproduce~\citep{nikishin2018improving, rlblogpost, henderson2018deep}.
This instability is attributed to multiple factors such as the random nature of exploration, the violation of the i.i.d assumption in data collection, the biased nature of value functions, and the general unpredictability of learning in deep neural networks~\citep{amin2021survey}. 
\citet{uc2023survey} overview methods and limitations for RL with LLMs in particular.

\textbf{\tractable Policies tend to be adversarially exploitable.} Even when learned policies are trained with a perfect reward signal, perform well at the task they are trained for, and generalize to a wide range of scenarios, they can still perform poorly in adversarial situations. This is a pressing concern, as models deployed into the real world can be adversarially attacked by humans or other AI systems. 
Even ``superhuman'' policies can fail catastrophically against policies specifically designed to exploit them \citep{Gleave2020Adversarial, wu2021adversarial, wang2022adversarial}. 
Adversarial policies can be found either by re-purposing existing deep-reinforcement learning algorithms or by manual human optimization in the case of prompt-injections and jailbreaks \citep{promptInjection2023, jailbreakChat2023, oneal2023, li2023multi, wolf2023fundamental, liu2023jailbreaking, rao2023tricking, wei2023jailbroken, shen2023anything} for language-models. 
Black-box access to a model (e.g., via API access) is sufficient for many adversarial policy attack algorithms, though white-box access (enabled for example by open-sourced or leaked model weights) enables even stronger exploits \citep{kos2017delving, casper2022white}.

\subsubsection{Policy Misgeneralization} \label{sec:policy_misgeneralization}

\textbf{\fundamental Policies can perform poorly in deployment even if rewards seen during training were perfectly correct.}
The deployment distribution can always differ from the training and evaluation distributions in real-world settings \citep{christiano2019worst}. 
Even with a correct reward signal, a policy can learn to competently pursue the wrong goal whenever the true goal is correlated with other events. 
\citet{shah2022goal, di2022goal} and \citet{hilton2020understanding} study this type of failure in-depth. 
\citet{shah2022goal} present an example scenario in which a systems trained with RLHF misgeneralizes to pursue the mechanism of reward administration itself instead of the intended goal.

\textbf{\fundamental Optimal RL agents tend to seek power.} 
RL agents have an incentive to seek power when possible to help them accomplish their goals \citep{turner2021seeking, Turner2019OptimalPT, Turner2022ParametricallyRD, ngo2022alignment, krakovna2023power, ngo2022alignment} 
Versions of this can emerge from the way that RLHF is typically used to finetune LLMs.
For example, a question-answering LLM trained with RLHF would be incentivized to influence human interlocutors in order to avoid conversations about challenging topics.
Sycophantic behavior from LLMs offers another example \citep{perez2022discovering}.

\subsubsection{Distributional Challenges}

There are challenges posed by the distribution of outputs produced by the model both before and after training.  

\textbf{\tractable The pretrained model introduces biases into policy optimization.} RLHF in LLMs typically begins with a base model that has been pretrained on internet text. 
This base model is typically used both as the initialization for the RL policy network and the reference model for KL-regularization.
\citet{korbak-etal-2022-rl} formalizes how RL with these KL penalties can be viewed as a form of Bayesian inference with the base model determining the prior.
While empirically useful, it causes the base model to significantly influence the final model. 
Using a base model that has been pretrained on web text is a convenient initialization -- not a principled one.
Moreover, internet text encodes harmful biases (e.g., about human demographics), which are then inherited by the downstream model~\citep{weidinger2021ethical}.
These biases can persist through RLHF training process.
For example, if sounding confident and producing correct answers are correlated in the base model, the reward model will learn that sounding confident is good and reinforce this in the policy.

\textbf{\tractable RL contributes to mode collapse.} RL finetuning decreases the diversity of samples produced by a model \citep{khalifa2021a, perez2022red, glaese2022improving, go2023aligning} (a phenomenon known as ``mode collapse''). 
\citet{openai2023gpt4} found that RLHF finetuning of GPT-4 harmed its calibration on question-answering.
\citet{santurkar2023whose} found LLMs finetuned with RLHF expressed a narrow distribution of political views.
Mode collapse is plausibly due in part to switching from the supervised pretraining objective to an RL objective \citep{song2023reward}. 
RL incentivizes the policy to output high-scoring completions with high probability, rather than with a probability in line with a training distribution.
Addressing this is complicated because mode collapse can be beneficial or harmful in different cases. 
For example, it is desirable if an LLM assistant is 90\% sure the answer to a question is ``yes'', it is better for the LLM to answer ``probably'' 100\% of the time rather than answering ``yes'' 90\% of the time and ``no'' 10\% of the time. 
On the other hand, some preferences are inherently distributional \citep{khalifa2021a,weidinger2021ethical} (e.g., gender balance).

\subsection{Challenges with Jointly Training the \blue{Reward Model} and \red{Policy}} \label{sec:joint_rm_policy_training} 

RLHF's dependence on training both a reward model and policy poses two unique problems. 

\textbf{\tractable Joint training induces distribution shifts.}
Learning both a reward model and a policy is technically challenging -- the reward model influences the learned policy, and the policy determines the distribution of the data used to train the reward. On one hand, if the reward model is trained on offline data, it is likely to misgeneralize \citep{levine2020offline}. On the other hand, if reward and policy are learned jointly by gathering feedback from policy samples, the system will be prone to ``auto-induced distributional shift'' \citep{krueger2020hidden, carroll22a}. Features with overestimated rewards will become gradually more present in the feedback data, and features with underestimated rewards will disappear.
Thus errors from the reward model can accumulate and become difficult to correct with feedback once the policy stops generating diverse alternatives \citep{wu2021recursively}.

\textbf{\tractable It is difficult to balance efficiency and avoiding overfitting by the policy.} The three key steps of RLHF can be performed synchronously, but in practice with LLMs, they are often performed serially. In this case, the reward model will typically be inaccurate off-distribution, which is precisely where the policy will learn to go \citep{gao2022scaling, levine2020offline}. This is usually solved by obtaining fresh preference labels after a certain number of iterations of policy training. Appropriately setting this hyperparameter is important. Too low and information in the preference labels is wasted; too high and the policy navigates to unreliable regions of the reward model \citep{mckinney2023fragility, christiano2017deep}. Without a labeled validation set in the regions the policy is exploring, it is difficult to detect reward over-optimization during training. 
Helpful approaches might include measuring KL-shift \citep{gao2022scaling} or tracking the amount of disagreement in an ensemble of reward models.

\section{Incorporating RLHF into a Broader Framework for Safer AI} \label{sec:rlhf_is_a_tool_in_the_toolbox_not_a_framework_for_safe_ai}

Because of the challenges surveyed in Section~\ref{sec:challenges_with_rlhf}, relying heavily on RLHF for developing safe AI poses risks. 
While RLHF is useful, it does not solve the fundamental challenges of developing human-aligned AI.
More generally, no single strategy should be treated as a comprehensive solution.
A better approach is defense in depth: multiple safety measures with uncorrelated failure modes. 
This is akin to assembling multiple layers of Swiss cheese---each has holes, but when layered can compensate for each other's failures \citep{hendrycks2021unsolved}.
While this type of approach is promising, it also comes with problems. For example, many of the challenges in Section~\ref{sec:challenges_with_rlhf} are not unique to RLHF, so it may be hard to find safety methods with uncorrelated failures. 
In this section, we discuss approaches that can be used to better \emph{understand} (\Cref{sec:understanding}), \emph{improve} on (\Cref{sec:improving}), and \emph{complement} (\Cref{sec:complementing}) RLHF in various ways as part of a broader agenda for AI safety.

\subsection{Frameworks for Better Understanding RLHF} \label{sec:understanding}

Although RLHF is becoming more widely used, there remain open questions about what factors are at play within it and how they influence the overall outcome. 
Here, we discuss approaches to address challenges for RLHF. 

\textbf{Psychology and human-computer interaction.}
Many of the open questions with RLHF involve the dynamics at play between humans and AI.
It remains a challenge to understand the conditions which best allow for safe, reliable human-computer interaction. 
Specifically, it is unclear what type of feedback (or combination thereof) is optimal for learning human goals, precisely how biases harm the quality of feedback, and how to best select and train human evaluators.
As discussed in \Cref{sec:challenges_with_rlhf}, human desires are difficult to express with a reward function \citep{skalsereward, bowling2023settling, vamplew2022scalar}.
Further work may be valuable toward inferring what beliefs humans are operating under and either asking for feedback while taking into account human uncertainty~\citep{biyik2019asking} or correcting for human biases \citep{reddy_where_2019, reddy_assisted_2020, chan_assistive_2019, tian_towards_2023}. 
Reward modeling systems must also take advantage of techniques that distinguish between humans with different levels of expertise~\citep{daniels2022expertise}, confidence~\citep{zhang2021confidence}, or noisiness~\citep{barnett2023active}.

\textbf{Sociology and social choice.}
AI alignment must address not only individuals' perspectives, but also the norms, expectations, and values of affected groups. 
Some works have begun to assess whether LLMs can be used to facilitate agreement between different humans \citep{bakker2022fine} and to codify the broad-ranging principles under which deployment of AI systems for public good can be assessed \citep{floridi2022unified, sartori2022sociotechnical}. 
The majority-rule problem with RLHF can also be improved by algorithms that explicitly model multiple evaluators~\citep{gordon2021disagreement, davani2022dealing, daniels2022expertise, gordon2022jury, barnett2023active}, that tune models to individuals~\citep{kumar2021designing}, or that use more sophisticated aggregation strategies~\citep{noothigattu2018voting}.
However, none of these approaches can solve the fundamental problem of how an AI system cannot be aligned to multiple groups of humans who hold conflicting viewpoints \citep{dobbe2021hard}. 
Many societies, however, confront this fundamental issue regularly.
For example, democracies seek to reflect social preferences by soliciting the feedback of individuals. 
These systems generally fail to align diverse preferences yet tend to be more acceptable than less-democratic alternatives. 
As such, it is important to analyze RLHF from the lens of social choice theory \citep{sen1986social} and work to understand whether the means by which it aggregates preferences is normatively acceptable.

\textbf{Assistance games.}
Assistance games, such as cooperative inverse RL (CIRL)~\citep{hadfield2016cooperative}, provide a framework to analyze algorithms like RLHF. They offer a mathematical model to evaluate different design decisions in the communication of preferences to learning systems. In an assistance game, a human and an agent act together in the environment. Both seek to optimize the human's latent reward function, while only the human can directly query this reward function. In this model, querying the human is simply an additional action that the robot can take, and it is possible to study different querying strategies or profiles. 
Studying RLHF as an assistance game emphasizes the performance of the human-robot team. 
This might suggest alternative preference elicitation methods.
Two examples are using active reward learning to determine when to collect feedback and which feedback to request first~\citep{sadigh2017active}, and leveraging dialogue models to learn desired feedback-seeking patterns~\citep{krasheninnikov2022assistance}. 
Of particular interest is understanding the consistency and convergence properties of RLHF, the impact of different error patterns from raters, and the effect of different rates of feedback.

\textbf{Bayesian inference.} Finetuning an LLM using RL with KL penalties on the differences between the pretrained model can be understood as a form of Bayesian inference: conditioning a prior (base LLM) on evidence about the desirable behavior of an LLM provided by the reward model \citep{korbak-etal-2022-rl}. 
This perspective on RLHF separates the modeling problem (defining a target distribution specifying the desired behavior of an LLM) and the inference problem (approximating that target distribution) \citep{korbak_2022, go2023aligning}. 
This can aid in answering questions about how the prior influences the outcome of RLHF. The typical target distribution of RLHF (a Boltzmann distribution) is a particular design choice and other distributions may address some of its limitations by, for example, differently fitting distributional preferences \citep{khalifa2021a}. 
Similarly, RLHF's inference algorithm (RL with KL penalties; equivalent to a variational inference approach \citep{korbak-etal-2022-rl}) could be replaced by a particular sampling strategy (e.g., rejection sampling or best-of-$n$ sampling).

\textbf{Worst-case behavior.} 
While RLHF seems to improve the average performance of a system, it is not clear what effects it has on worst-case behavior. 
It was not designed to make systems adversarially robust, and empirical vulnerabilities of systems trained with RLHF have been demonstrated with jailbreaks and prompt injection attacks \citep{promptInjection2023, jailbreakChat2023, oneal2023, li2023multi, wolf2023fundamental, liu2023jailbreaking, rao2023tricking, wei2023jailbroken, shen2023anything}.
As a consequence, it would be valuable to better understand the worst-case behaviors of RLHF systems, potentially through the lenses of theoretical properties \citep{wolf2023fundamental, el2022sok}, decision theory \citep{casper2020achilles}, adversarial attacks \citep{perez2022red, perez2022discovering,casper2023explore, ziegler2022adversarial, carlini2023aligned}, or rigorous evaluations \citep{arcevals, openai2023gpt4, shevlane2023model}.

\subsection{Addressing Challenges with RLHF} \label{sec:improving}

\begin{figure}
    \centering
    \includegraphics[width=\textwidth]{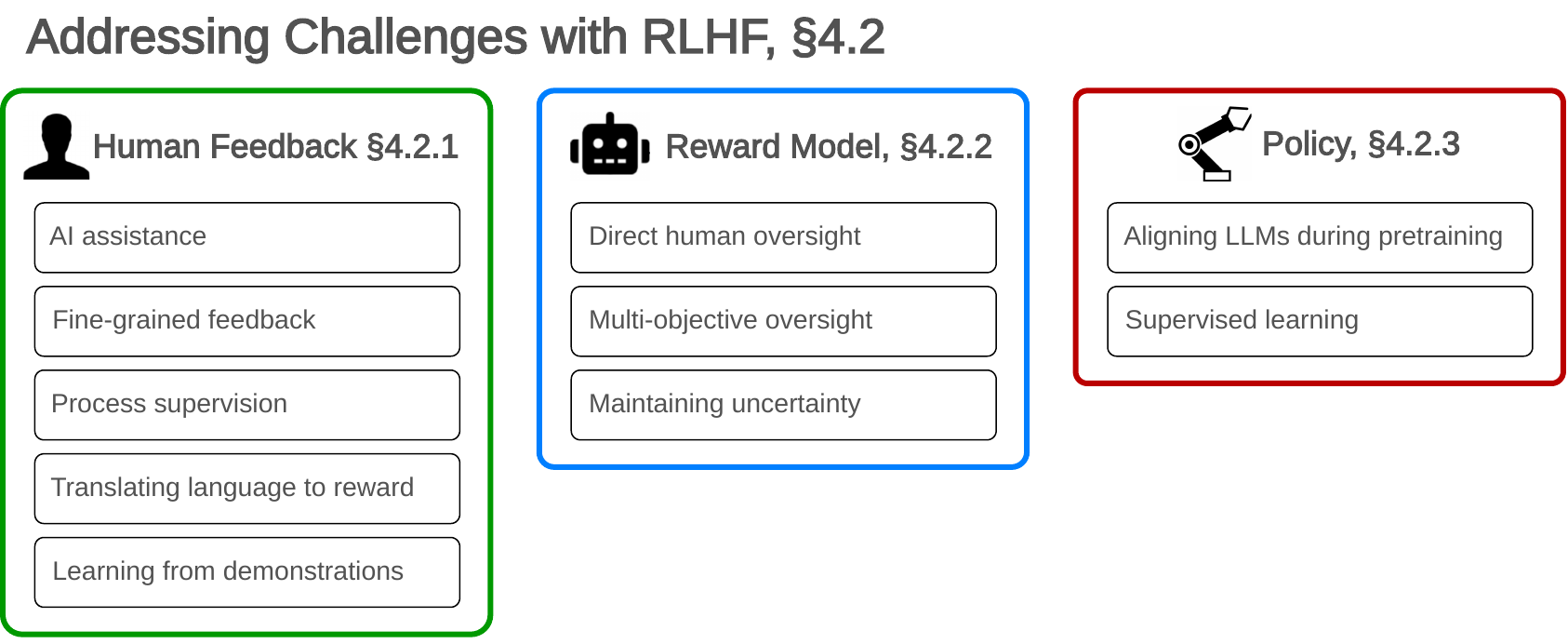}
    \caption{\textbf{Strategies that can be used to address various problems with RLHF.} Each approach is discussed in \Cref{sec:improving}.
}
    \label{fig:improving}
\end{figure}

Just as RLHF has challenges involving feedback (\Cref{sec:feedback}), the reward model (\Cref{sec:reward_model}), and the policy (\Cref{sec:policy}), there are various methods that can replace or combine with parts of the RLHF pipeline to address each of these types of challenges. \Cref{fig:improving} outlines these methods. 
See also \citet{wang2023aligning} for a survey of methods for aligning LLMs. 

\subsubsection{Addressing Challenges with \green{Human Feedback}} \label{sec:addressing_human_feedback}

\textbf{Providing feedback with AI assistance.} One way to amplify the abilities of humans is to have AI tools assist in generating feedback. 
Engineering prompts for an AI system and using it to automate feedback can substantially increase practicality and cost-effectiveness due to reduced reliance on humans.
Nonetheless, AI-generated feedback still fundamentally depends on humans because (1) the models providing feedback are trained on human-generated data, and (2) humans control prompts and the process of incorporating feedback.
There are several notable examples of AI-generated language feedback \citep{bai2022constitutional, saunders2022self, ye2023selfee, kim2023aligning, akyurek2023rl4f, madaan2023self, chen2023improving, gilardi2023chatgpt, lee2023rlaif} with research agendas like Recursive Reward Modeling \citep{leike2018scalable} and AI Safety via debate \citep{irving2018ai, du2023improving}.
However, AI-generated feedback has drawbacks. Humans often disagree with AI feedback. 
The rate of human/AI disagreement will vary by task, but \citet{perez2022discovering}, \citet{casper2023explore}, and \citet{lee2023rlaif} found this to happen up to 10\%, 46\%, and 22\% of the time respectively in different experiments. 
Machines can also exhibit correlated failure modes not found in humans, such as vulnerabilities to some adversarial attacks. The extent to which AI feedback is a viable way to safely augment human feedback remains uncertain. 
However, it cannot theoretically be a comprehensive solution to AI alignment due to the bootstrapping problem behind ensuring the feedback-providing AI is aligned.

\textbf{Fine-grained feedback.} 
Many problems with feedback involve difficulty conveying precise information via the feedback signal (\Cref{sec:feedback_type}).
To address this, \citet{wu2023finegrained} and \citet{cabi2019scaling} use feedback on specific portions of examples and \citet{wu2023finegrained} use feedback with respect to different goals of the model (e.g., correctness, relevance).
This might improve the quality of the learned reward models at the cost of human feedback being more expensive to provide. 
Fine-grained feedback is not yet well studied nor widely adopted, so additional work to understand its advantages and feasibility will be valuable.

\textbf{Process-based supervision.} 
One challenge with training AI systems to solve problems is the difficulty of supervising performance on multi-step procedures. 
In RL, rewards can be very sparse for such problems. 
To address this, some works have trained LLMs to better solve multi-step math problems with process supervision \citep{uesato2022solving, lightman2023let}.

\textbf{Translating natural language specifications into a reward model.} 
Many issues with RLHF arise due to the difficulty of fitting a reward function using some constrained type of feedback. 
An alternative approach can be to generate a reward signal more directly from natural language directions, bypassing the need for feedback on examples. 
This approach could resemble a technique used by \citet{bai2022constitutional} which involved using prompts to guide an AI assistant to identify responses that violated certain user-defined specifications. 
Moreover, \citet{luketina2019survey} surveys other possible techniques to accomplish this goal in non-LLM settings.

\textbf{Learning rewards from demonstrations.} An alternative approach to learning a reward model, known as inverse reinforcement learning (IRL) \citep{ng2000algorithms, ramachandran2007bayesian, ziebart2008maximum}, involves humans providing demonstrations instead of offering feedback on ones generated by the model. \citet{jeon2020reward} and \citet{biyik2022learning} propose systematic ways of combining demonstrations, preferences, and possibly other types of human feedback to learn reward functions. While demonstrations carry rich information and avoid the need to have a system learn from its own generations, they are often more difficult to gather because they require higher effort and expertise to perform the task. 
Additionally, the quality of demonstrations is limited by the talent of whatever expert is providing them, which warrants more research on learning from suboptimal human demonstrations (e.g., \citet{brown2019extrapolating, zhang2021confidence}).

\subsubsection{Addressing Challenges with the \blue{Reward Model}} \label{sec:addressing_reward_model}

\textbf{Using direct human oversight.} Although learning a reward model is efficient, it might be necessary to directly provide rewards \citep{macglashan2017interactive} for RL training in certain safety-critical situations.

\textbf{Multi-objective oversight.} Richer multi-objective signals that rate outputs on multiple objectives~\citep{vamplew2022scalar} could lead to more flexible oversight. 
Current reward models assume that expert feedback is drawn from an underlying unimodal reward function~\citep{barnett2023active, myers2022learning}. 
But this is overly simplistic \citep{skalsereward, bowling2023settling}. 
For instance, it can lead to a reward model that merely captures the preferences of the majority, and suppresses the preferences of minorities as noise. 
Using constraints~\citep{ malik2021inverse, lindner2023learning} or reward models that account for the diversity of preferences by assuming underlying reward functions to be multimodal~\citep{myers2022learning, bakker2022fine, barnett2023active, siddique2023fairness, bhatia2020preference} can help mitigate this issue. 
Multi-objective oversight can also be useful for steering systems toward desired balances between competing values (e.g., helpfulness and harmlessness).

\textbf{Maintaining uncertainty over the learned reward function.} Given the challenges of accurately learning the appropriate reward function, several studies have emphasized the importance of taking uncertainty in the learned functions into account. \citet{yue2023clare} and \citet{liangreward} tackle this by having the policy avoid types of states unseen by the reward model.
Using an ensemble of reward functions has also been used to address these challenges \citep{christiano2017deep}, demonstrating that this approach can enhance the diversity of text output \citep{rame2023rewarded} and its applicability for active learning \citep{gleave2022uncertainty}.
Other strategies can include forms of risk-aversion \citep{hadfield2017inverse} or handling uncertainty with a safe ``shield'' policy \citep{jansen2018safe, srinivasan2020learning, cohen2020curiosity}.

\subsubsection{Addressing Challenges with the \red{Policy}} \label{sec:addressing_policy}

\textbf{Aligning LLMs during pretraining.} RLHF in LLMs typically begins by pretraining the LLM on internet text which includes a large amount of undesirable content. \citet{korbak2023pretraining} argue that it can be more effective to use human feedback during pretraining by using a reward model to filter, weight, or annotate pretraining data. This also simplifies the process of aligning models by having them exhibit desirable behaviors from the outset rather than having them learn undesirable behavior and then attempt to unlearn it during finetuning.

\textbf{Aligning LLMs through supervised learning.} 
Several techniques for aligning LLMs with human preferences obtain results competitive with RLHF by using supervised learning to complement \citep{ramamurthy2022reinforcement} or replace RL. 
The simplest variant of this is to perform standard supervised learning on well-curated data. 
Curation can involve filtering out bad demonstrations \citep{gehman-etal-2020-realtoxicityprompts, weibl2021, dong2023raft}, compiling a small set of good demonstrations \citep{palms,sanh_t0, ibarz2018reward, stiennon2020learning, chung2022_scaling_instruction, biyik2022learning, zhou2023lima},
or generating good demonstrations using an LLM, e.g., after conditioning human feedback provided in natural language \citep{scheurer2022training, scheurer2023training, chen2023improving, xu2023shattering}.
A different family of methods augments the language modeling objective to utilize feedback provided by the reward model \citep{korbak2023pretraining, yuan2023rrhf, rafailov2023direct}. This last setting shares similarities with offline RL, which focuses on training an optimal policy using demonstrations annotated with rewards \citep{levine2020offline, snell_ilql, hu2023aligning}.

\subsection{RLHF is Not All You Need: Complementary Strategies for Safety} \label{sec:complementing}

Other technical approaches to AI safety should be studied and implemented alongside RLHF.
Establishing trust with AI systems should be approached with a combination of principled design choices, rigorous testing, interpretability, verification, and theoretical guarantees where possible \citep{leike2018scalable}.
See also \citet{critch2020ai}, \citet{hubinger2020overview}, \citet{hendrycks2021unsolved}, and \citet{ngo2022alignment} for additional overviews of strategies for building safer AI.

\textbf{Robustness.} As discussed in \Cref{sec:policy}, models trained with RLHF can still exhibit undesired behavior due to distributional shifts between training and deployment. For example, adversarially engineered user inputs cause an LLM to output harmful text. To mitigate this problem, developers should use tools to generate inputs which result in undesired behavior and train against these adversarial examples \citep{zhang2019adversarial, ziegler2022adversarial, perez2022red, casper2023explore}. Anomaly detection techniques \citep{omar2013machine} can also be useful for flagging abnormal inputs likely to trigger bad behavior.
Ensuring the security of important AI training runs against malicious human evaluators and/or outside cybersecurity threats will also be valuable. 

\textbf{Risk assessment and auditing.}
Although training processes should be crafted to produce models that are safe by design, evaluations are another layer of defense.
Passing an evaluation is not proof of safety, but as is the case in almost every safety-critical industry, rigorous evaluations of capabilities and risks helps to spot hazards and establish trust.
In practice, this should involve both in-house and second-party evaluations \citep{openai2023gpt4, arcevals, perez2022discovering}.
As with adversarial training for robustness, the development of improved red teaming techniques will be important \citep{perez2022red, casper2023explore}. 

\textbf{Interpretability and model editing.} 
Generating human-understandable explanations for the behavior of AI systems is currently an unsolved problem. 
Progress in explainability and interpretability could help verify hypotheses about how models make decisions \citep{geiger2023causal}, including whether the decision-making process is trustworthy. 
In this way, it could be possible to gain confidence that models will (or will not) behave in a safe way without necessarily conducting extensive testing of the models \citep{jacovi2021formalizing}. 
Red-teaming can also be complemented by interpretability techniques \citep{rastogi2023supporting, rauker2023toward}, especially for purposes of identifying adversarial inputs \citep{ziegler2022adversarial, casper2023robust, casper2023benchmarking} or anomalous inputs \citep{pang2021deep}. 
In another direction, better understanding the internal mechanisms of models can aid in directly editing model weights or intervening on internal activations in order to improve truthfulness \citep{li2023inferencetime}, modify a model's factual knowledge \citep{meng2023locating, meng2022mass, hernandez2023measuring, hase2023does}, or otherwise steer model behavior \citep{cui2022local}.

\section{Governance and Transparency} \label{sec:governance_and_transparency}

Social scientists and policymakers have increasingly focused on the need for governance frameworks to develop and deploy AI systems responsibly. 
Across historical contexts, a hallmark of mature scientific fields is the open sharing of research findings \citep{shapin2011leviathan} to allow experts to understand progress \citep{gilbert2021subjectifying}.
Below we overview components of an RLHF governance agenda, including outstanding questions and risk dimensions.

\textbf{Incentives and requirements for safety.} 
Competition between labs can generate harmful race dynamics \citep{dafoe2018ai} because of tradeoffs between competitiveness and caution. 
This suggests a role for governance in promoting a healthier environment for safe AI research, development, and deployment \citep{dafoe2018ai, perry2019ai, falco2021governing, cihon2019standards, anderljung2023frontier}.
Governance in this form could involve mandates for independent auditing, evaluations, and certification \citep{shavit2023does, mokander2023auditing, arcevals, hadfield2023regulatory, shevlane2023model}; monitoring for post-deployment problems \citep{hendrycks2016baseline}; influence over resources including hardware and data \citep{brief2020ai, chan2023reclaiming}; and prohibiting deployment unless critical standards are met, as in the case of the U.S. Food and Drug Administration's oversight of clinical trials for testing potential new treatments \citep{junod2008fda}. 

\textbf{Transparency and auditing.} 
A sustained commitment to transparency would make the existing RLHF research environment more robust from a safety standpoint.
First, the disclosure of some details behind large RLHF training runs would clarify a given organization's norms for model scrutiny and safety checks. 
Second, increased transparency about known efforts to mitigate risks could improve safety incentives and suggest methods for external stakeholders to hold companies accountable.
Third, and most relevant for the present paper, transparency would improve the AI safety community's understanding of RLHF and support the ability to track technical progress on its challenges. 
Some level of disclosure is a precondition to evaluate the viability of the technical RLHF safety agenda over time and allow for community contribution to it. 
For all of these reasons, working to incorporate transparency standards into an AI governance framework will be important \citep{larsson2020transparency, anderljung2023frontier}.
It is possible that public disclosure of details critical to the development of model capabilities might lead to the unwanted proliferation of AI technologies that could be misused.
However, detailing safety measures will often not require divulging implementable details, and when it does, private disclosure to second-party auditors \citep{mokander2023auditing, arcevals, hadfield2023regulatory, shevlane2023model} offers a solution.

\begin{figure}
    \centering
    \includegraphics[width=\textwidth]{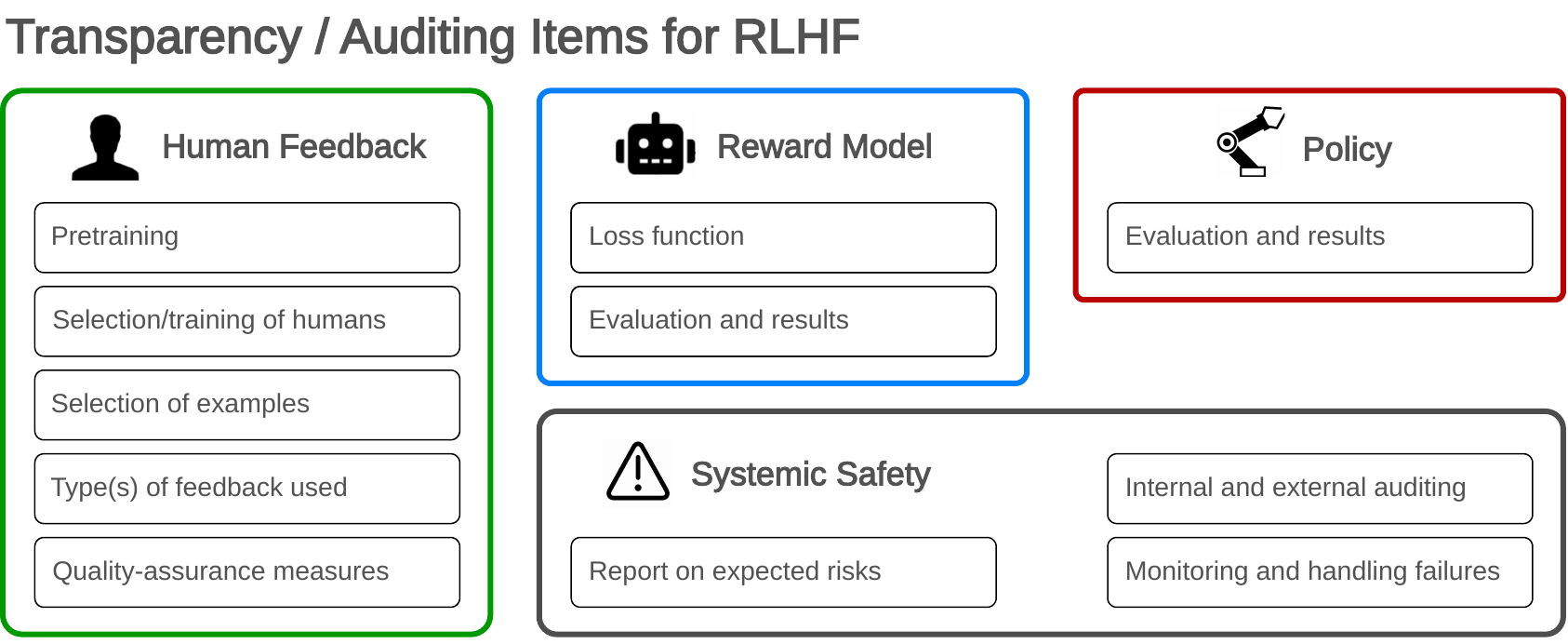}
    \caption{\textbf{Details behind an implementation of RLHF that, if disclosed, could be indicative of risks.} See \Cref{sec:governance_and_transparency} for a complete discussion. Companies using RLHF to train models for high-stakes or safety-critical applications should maintain transparency with the public and/or auditors about key details of their approach. 
}
    \label{fig:transparency}
\end{figure}

As more specific policy prescriptions are beyond our scope, we encourage elaboration on these topics as part of a future research agenda. 
Below, however, we outline specific types of details that, if disclosed, could be indicative of risks and should be accounted for when auditing AI systems developed using RLHF. See also \Cref{fig:transparency}.

\hspace{0.2in} \green{Human feedback} details:
\begin{itemize}
    \item \textbf{A description of the pretraining process including details about what data was used} to make apparent possible biases that pretraining can cause.
    \item \textbf{How human evaluators were selected and trained} to provide information about risks of evaluators being malicious, unrepresentative, or incapable.
    \item \textbf{The process by which examples were selected to obtain feedback} to invite scrutiny about their representativeness and whether sufficient adversarial training was used. If examples were crowdsourced from a publicly-available application, details about what measures were taken to avoid data poisoning attacks should be provided. 
    \item \textbf{The type(s) of human feedback used} (e.g., binary comparisons, scalar feedback, etc.) to suggest what risks might be caused by insufficiently abundant or rich feedback.
    \item \textbf{A report on measures taken for quality assurance in feedback collection and inter-rater consistency} to ensure that effective quality control measures were taken.
\end{itemize}

\hspace{0.2in} \blue{Reward model} details:
\begin{itemize}
    \item \textbf{The loss function used to fit the reward model and how disagreement was modeled} (e.g., as noise) to help with analyzing the degree of misspecification when fitting the reward model.
    \item \textbf{A report on reward model evaluation and results} to suggest possible problems from a misaligned reward model. The evaluation should involve red teaming.
\end{itemize}  
    
\hspace{0.2in} \red{Policy} details:
\begin{itemize}
    \item \textbf{A report on policy evaluation and results} to suggest possible troubles from a misaligned policy. The evaluation should involve red teaming and include assessment for risky capabilities (e.g., the ability to deceive a human). 
\end{itemize}
    
\hspace{0.2in} Systemic safety measures
\begin{itemize}
    \item \textbf{A report on internal and external audits and red teaming} to ensure accountability and disclose risks that are identified. 
    \item \textbf{A report on expected risks and anticipated failure modes} to ensure accountability.  
    \item \textbf{Plans for monitoring and correcting failures that emerge} to support post-deployment safety. 

\end{itemize}

How these types of risks should be documented remains an area of active work in AI governance. 
Similar questions have been asked in an investigation by the US Federal Trade Commission into OpenAI \citep{ftc2023} but in response to problems with ChatGPT rather than proactively. 
Salient documentation proposals focus on regular reporting of reward components \citep{gilbert2022reward} and the ability to compare the capabilities of language models according to standard benchmarks \citep{liang2022holistic}. For the longer term, incorporating beneficial standards for safety and transparency into norms and regulations affecting AI is an ongoing challenge. 

\textbf{Concerns for social and economic equity.} Although this paper has focused on technical challenges with RLHF, there are social and economic ones as well which governance and industry should work to address. 
For example, OpenAI has paid Kenyan knowledge workers at a rate of less than \$2 USD per hour \citep{timeExclusiveHour} for work which was mentally and emotionally demanding \citep{hao2023hidden}. 
Human subjects used in RLHF research should not be systematically selected simply for their availability or low cost \citep{united1978belmont}.
Costs, benefits, and influence over RLHF models should be equitably distributed across different communities \citep{whittlestone2021societal, eloundou2023gpts}.
There is an additional possibility that powerful AI systems will be highly profitable and serve to concentrate large amounts of wealth and power into the hands of a few \citep{o2020windfall, Chan2023HarmsFI}.
Thus, policies that address inequalities and protect vulnerable populations (e.g. impacted communities, whistleblowers) will be increasingly important.

\section{Discussion} \label{sec:conclusion}

\textbf{While some problems with RLHF are \violet{tractable}, others are \orange{fundamental}.} 
Technical progress in some respects is tractable, and this room for progress should be seen as a cause for concerted work and optimism. 
Even some of the fundamental problems that we overview can be alleviated with improved methodology even though they cannot be fully solved by RLHF.
However, the fundamental nature of these problems requires that they be avoided or compensated for with non-RLHF approaches.
Hence, we emphasize the importance of two strategies: (1) evaluating technical progress in light of the fundamental limitations of RLHF and other methods, and (2) addressing the sociotechnical challenges of aligning to human values by committing to both defense-in-depth safety measures and openly sharing research findings with the wider scientific community.

\textbf{RLHF = Rehashing Lessons from Historical Failures?}
RLHF offers new capabilities but faces many old problems.
Its use by \citeauthor{christiano2017deep} dates to 2017, and the individual components of it (preference elicitation, fitting a reward model, and policy optimization) have a history of technical and fundamental challenges in the fields of human-computer interaction and AI safety. 
In 2023, RLHF was \href{https://www.alignmentforum.org/posts/vwu4kegAEZTBtpT6p/thoughts-on-the-impact-of-rlhf-research#The_case_for_a_positive_impact:~:text=I%20think%20it%20is%20hard%20to%20productively%20work%20on%20more%20challenging%20alignment%20problems%20without%20first%20implementing%20basic%20solutions.}{described} by the first author of \citet{christiano2017deep} as a ``basic solution'' intended to make it easier to ``productively work on more challenging alignment problems'' \citep{christiano2023thoughts}.\footnote{\citet{christiano2023thoughts} mentions debate \citep{irving2018ai} and recursive reward modeling \citep{leike2018scalable} as examples of `more challenging alignment problems.' See also an outline of proposals in \citet{hubinger2020overview}.}
Some challenges and questions that we have covered are rather unique to RLHF such as ones involving jointly training the reward model and policy (\Cref{sec:joint_rm_policy_training}).
However, many other problems are instances of broader ones in machine learning such as challenges with RL policies (\Cref{sec:policy}).
Others still are fundamental problems with AI alignment such as determining whose values are encoded into AI in a diverse society of humans (\Cref{sec:misspecification}).
The successes of RLHF should not obfuscate its limitations or gaps between the framework under which it is studied and real-world applications (see \Cref{app:improved_model}). 
An approach to AI alignment that relies on RLHF without additional techniques for safety risks doubling-down on flawed approaches to AI alignment. 
Thus, it will be important to continue working to better understand RLHF while respecting its limitations.

\textbf{Moving forward.}
RLHF has clear advantages for aligning AI systems with human goals.
As a result, it has been key to the development of state-of-the-art LLMs and will likely continue to play a major role in modern AI. 
However, its use and influence should be accompanied by a commensurate research effort to better understand RLHF and address its flaws. 
Because it optimizes for human approval, RLHF in particular demands a special type of caution because many of its failures will actively tend to be ones that humans struggle to notice. 
It will be important to approach RLHF cautiously and work to incorporate it into a more holistic framework \citep{khlaaf2023toward} for safer AI with multiple layers of protection from failures \citep{hendrycks2021unsolved}. 
Because some of the challenges with RLHF are fundamental to the AI alignment problem itself, moving forward will require confronting the basic choices and assumptions behind any given approach to aligning AI and who controls it \citep{dobbe2021hard}. 
Moving forward, we urge that those working to develop advanced LLMs using RLHF both contribute toward resolving its open challenges and maintain transparency about the details of their approach to safety and any anticipated risks.

\section*{Contributions}

Stephen Casper and Xander Davies served as the central writers and organizers. 

Claudia Shi, Thomas Krendl Gilbert, Jérémy Scheurer, Javier Rando, Rachel Freedman, Tomasz Korbak, David Lindner, Pedro Freire, Tony Wang, Samuel Marks, Charbel-Raphaël Segerie, Micah Carroll, Andi Peng, Phillip Christoffersen, Mehul Damani, Stewart Slocum, Usman Anwar, Anand Siththaranjan, Max Nadeau, Eric J. Michaud, Jacob Pfau, Xin Chen, Dmitrii Krasheninnikov, Lauro Langosco, and Peter Hase contributed to writing and planning the paper. 

Erdem B{\i}y{\i}k, Anca Dragan, David Krueger, Dorsa Sadigh, and Dylan Hadfield-Menell served as advisors.

\section*{Acknowledgements}
We thank Sam Bowman, Adam Jermyn, Ethan Perez, Alan Chan, Gabriel Recchia, Robert Kirk, and Nathan Lambert for their helpful feedback. This work was facilitated in part by the Harvard AI Safety Team and MIT AI Alignment.

\newpage

\bibliographystyle{plainnat}
\bibliography{bibliography}

\newpage

\appendix

\section{An Improved Model of the Human Feedback Process} \label{app:improved_model}

As illustrated in \Cref{eq:feedback}, the feedback process in RLHF is typically modeled with a single human $\mathcal{H}$ with internal reward function $r_{\mathcal{H}}$; examples sampled from the base model: $x_i \sim \pi_\theta$; and feedback as a function of the human, example, and noise: $y_{i} = f(h, x_i, \epsilon_{i})$. However, as discussed in \Cref{sec:challenges_with_rlhf}, this is a misspecified model of the process: there is not a single human, humans values are not representable with a reward function, human actions are dependent on context, and the sampling process can involve a human. Thus we propose an alternative formulation.

Let $\Delta \mathcal{H}$ refer to a joint distribution of humans (or groups thereof if feedback is provided collaboratively) used for obtaining samples and feedback denoted as $\mathcal{H}_{j}^{\textrm{sample}}$ and $\mathcal{H}_{j}^{\textrm{feedback}}$. 
A dataset of examples is sampled from $\pi_\theta$ (or some other source) where each example $x_i$ is defined to be a batch of one or more generations from the base model. Importantly, $x_i$ may not contain all information about the world state (e.g., if $x_i$ is a 2D rendering of a 3D environment), and the human may be able to observe more than just the model's output (e.g., if interpretability tools are used to aid in evaluation). So let $v$ be a rendering function that maps $\pi_\theta$ and $x_i$ to what a human sees.
The behavior of humans varies over time and in different contexts, so let $c_{i}^{\textrm{sample}}$ and $c_{i}^{\textrm{feedback}}$ represent particular contexts for sampling and feedback collection.
Denote the sampling process as $s$ which maps the base model $\pi_\theta$, a human $\mathcal{H}_{j}^{\textrm{sample}}$, and context $c_{i}^{\textrm{sample}}$ to some example $x_i$. Notably, $s$ could ignore the base model and generate offline samples from some other source.
Finally, let $f$ map a human $\mathcal{H}_{j}^{\textrm{feedback}}$, rendered example $v(\pi_\theta, x_i)$, and context $c_{i}^{\textrm{feedback}}$ to feedback $y_i$.
The data collection process can thus be more completely modeled as:
\begin{align}
\mathcal{H}_{j}^{\textrm{sample}}, \mathcal{H}_{j}^{\textrm{feedback}} \sim \Delta\mathcal{H}, \hspace{2em} x_i \sim s(\pi_\theta, \mathcal{H}_{j}^{\textrm{sample}}, c_{i}^{\textrm{sample}}), \hspace{2em}  y_{i} = f(v(\pi_\theta, x_i), \mathcal{H}_{j}^{\textrm{feedback}}, c_{i}^{\textrm{feedback}})\label{eq:improved_feedback}
\end{align}
which highlights a need for future work to better account for the aspects of this process that are commonly not accounted for when training systems with RLHF. 

\section{Rationale for Why Challenges Were Categorized as Tractable or Fundamental} \label{app:explanations}

In \Cref{sec:challenges_with_rlhf}, we categorize problems as \violet{tractable} or \orange{fundamental}. The key distinction between the two is that fundamental challenges are substantial enough that overcoming them would require a method that is no longer a form of RLHF.
Although many of the fundamental problems we identify can be alleviated by improving how RLHF is approached, they could be fully addressed with RLHF. As a result, they should be either avoided by not using RLHF or compensated for by other safety measures.
This distinction is soft, and some categories of challenges are marginal. Here, we briefly explain each categorization. 

\subsection{Problems from \Cref{sec:feedback}:}

\textbf{\tractable Selecting representative humans and getting them to provide quality feedback is difficult:} This can be addressed by studying and improving the selection and training of evaluators.

\textbf{\tractable Some evaluators have harmful biases and opinions:} This can be addressed by studying and improving the selection and training of evaluators.

\textbf{\tractable Individual human evaluators can poison data:} This can be addressed with improved evaluator selection and quality assurance measures.  

\textbf{\tractable Humans make simple mistakes due to limited time, attention, or care:} This is marginal because human mistakes can never fully be overcome. However, they can be addressed with improved working conditions and quality assurance procedures.  

\textbf{\tractable Partial observability limits human evaluators:} Human evaluators can be provided with all information available in the policy's observations (although representing this in an easily-comprehensible way may be challenging). 

\textbf{\fundamental Humans cannot evaluate performance on difficult tasks well:} Human intelligence and cognitive capacity are limited. Humans cannot be expected to properly evaluate the performance of superhuman models on complex tasks. Thus, solving this problem would require no longer using human feedback in the way that RLHF does. 

\textbf{\fundamental Humans can be misled, so their evaluations can be gamed:} Human fallibility cannot fully be overcome, especially against optimization pressure from the learned policy.

\textbf{\tractable Data collection can introduce harmful biases:} This can be addressed with improved data curation. 

\textbf{\fundamental There is an inherent cost/quality tradeoff when collecting human feedback:} This tradeoff is unavoidable in practice -- obtaining diverse and high-quality examples (e.g. from long chatbot conversations) requires more effort.

\textbf{\fundamental RLHF suffers from a tradeoff between the richness and efficiency of feedback types:} This tradeoff is unavoidable for data collection in practice -- richer annotations require more effort. 

\subsection{Problems from \Cref{sec:reward_model}:}

\textbf{\fundamental An individual human's values are difficult to represent with a reward function:} This problem is marginal. It can be improved in practice by improved modeling, but RLHF-based solutions will be limited by the intractability of perfectly modeling context and troubles with the reward hypothesis \citep{skalsereward, bowling2023settling}.

\textbf{\fundamental A single reward function cannot represent a diverse society of humans:} Trivial. Instead of being a fundamental limitation with RLHF, this is a broader limitation of AI alignment itself. 

\textbf{\fundamental Reward models can misgeneralize to be poor reward proxies, even from correctly-labeled training data:} This problem is marginal because it can and should be addressed by improved sampling in practice. However, it is impossible to perfectly represent a distribution with infinite support from a finite sample. Additionally, the deployment distribution will always differ from the training and evaluation distributions in real-world settings \citep{christiano2019worst}.

\textbf{\fundamental Optimizing for an imperfect reward proxy leads to reward hacking:} If a reward model is imperfect, reward hacking will always be a possibility from RL. 

\textbf{\tractable Evaluating reward models is difficult and expensive:} This can be addressed by performing thorough and expensive evaluations.  

\subsection{Problems from \Cref{sec:policy}:}

\textbf{\tractable It is (still) challenging to optimize policies effectively:} This can be addressed with advancements in RL methodology.

\textbf{\tractable Policies tend to be adversarially exploitable:} This problem is marginal because achieving certified adversarial robustness against practical threat models has empirically been intractable. Nonetheless, this can be addressed with robust optimization techniques. 

\textbf{\fundamental Policies can perform poorly in deployment even if rewards seen during training were perfectly correct:} This problem is marginal because it can and should be addressed by improved sampling in practice. However, it is impossible to perfectly represent a distribution with infinite support from a finite sample. Additionally, the deployment distribution will always differ from the training and evaluation distributions in real-world settings \cite{christiano2019worst}. 

\textbf{\fundamental Optimal RL agents tend to seek power:} Power is instrumentally useful for agents. 

\textbf{\tractable The pretrained model introduces biases into policy optimization:} This can be addressed with improved base models.

\textbf{\tractable RL contributes to mode collapse:} This can be addressed with forms of RL that optimize for distribution-matching in desired instances.

\subsection{Problems from \Cref{sec:joint_rm_policy_training}:}

\textbf{\tractable Joint training induces distribution shifts:} This can be mitigated with synchronous learning or other strategies. 

\textbf{\tractable It is difficult to balance efficiency and avoiding overfitting by the policy:} This can be addressed with improved training methodology.

\end{document}